\documentclass[lettersize,journal]{IEEEtran}
\usepackage{enumitem}
\usepackage{amsmath,amsfonts}
\usepackage{algorithmic}
\usepackage{algorithm}
\usepackage{array}
\usepackage[caption=false,font=normalsize,labelfont=sf,textfont=sf]{subfig}
\usepackage{textcomp}
\usepackage{stfloats}
\usepackage{url}
\usepackage{tabularx}
\usepackage{verbatim}
\usepackage{graphicx}

\usepackage{cite}
\usepackage{xcolor}
\usepackage{booktabs}
\usepackage{multirow}
\usepackage{rotating}    
\usepackage{amsmath} 
\usepackage[normalem]{ulem} 
\usepackage{caption} 
\usepackage[table]{xcolor}
\usepackage{tabularx, booktabs}
\usepackage{graphicx, multirow, booktabs, rotating}

\begin{document}
\title{Channel-Level Semantic Perturbations: Unlearnable Examples for Diverse Training Paradigms}

\author{Bo Wang, Jia Ni, Mengnan Zhao, Zhan Qin, Kui Ren,~\IEEEmembership{Fellow,~IEEE}
\thanks{Manuscript received April 16, 2026.}

\thanks{Bo Wang and Jia Ni are with the School of Information and Communication Engineering, Dalian University of Technology, Dalian 116024, China.}

\thanks{Mengnan Zhao is with the School of Computer Science and Technology, Anhui University, Hefei 230601, China (e-mail: zmn@ahu.edu.cn).}

\thanks{Zhan Qin and Kui Ren are with the School of Computer Science and Technology, Zhejiang University, Hangzhou 310058, China.}

}

\markboth{IEEE Transactions on Dependable and Secure Computing}%
{Wang \MakeLowercase{\textit{et al.}}: Channel-Level Semantic Perturbations: Unlearnable Examples for Diverse Training Paradigms}


\maketitle


\begin{abstract}

The unauthorized use of personal data in model training has emerged as a growing privacy threat. Unlearnable examples (UEs) address this issue by embedding imperceptible perturbations into benign examples to obstruct feature learning. However, existing studies mainly evaluate UEs under from-scratch training settings, leaving their behavior under the widely adopted pretraining–finetuning (PF) paradigm largely unexplored. In this work, we provide the first systematic investigation of unlearnable examples across diverse training paradigms. Our analysis reveals that loading and freezing pretrained weights significantly weakens the effectiveness of existing UEs methods.
We further explain these findings through semantic filtering: while UEs tend to induce models to overfit non-semantic noise, thereby weakening their semantic extraction capabilities, under the PF paradigm, frozen shallow layers preserve data semantics, effectively filtering out distracting information like unlearnable noise.
Guided by these insights, we propose a hierarchical deception strategy, Shallow Semantic Camouflage (SSC), that confines the generation process to a semantically valid subspace, aiming to bypass the semantic suppression introduced by pretrained weights. Extensive experiments demonstrate that our method consistently preserves data unlearnability even under challenging training paradigms, such as shallow-layer freezing and semantic-focused pretraining (SF-Pretrain), bridging the critical gap in pretrain-based unlearnable learning.

\end{abstract}


\section{Introduction}


\IEEEPARstart{T}{he} availability of large-scale datasets \cite{imagenet,coco} has driven rapid progress in deep learning.
Such data is frequently scraped from public sources and lacks effective access restrictions~\cite{Hong2025A,Krotov2022Big}, leading to growing concerns about the unauthorized use of data and intellectual property protection~\cite{dataprotection,somepalli2023diffusion,10458329}.
For example, massive web-collected datasets frequently contain personal or sensitive user information, and such information can be memorized during model optimization \cite{ Yu2023Bag, Lee2021Deduplicating,9633176}, posing substantial privacy-leakage risks~\cite{Ishihara2023Training}. More seriously, prior work~\cite{Pang2024ReconstructionOD, Lukas2023Analyzing, Meisenbacher2025The} has demonstrated that even with data sanitization, adversaries can still reconstruct high-fidelity private content or infer sensitive attributes from trained models. To mitigate these issues, recent research~\cite{11316260, DH,wang2024provably, MEM,9186317} has proposed the unlearnable technique, which reduces model generalization on private data and thereby discourages unauthorized data exploitation.
\begin{figure}[t]
    \centering
    \includegraphics[width=0.97\linewidth]{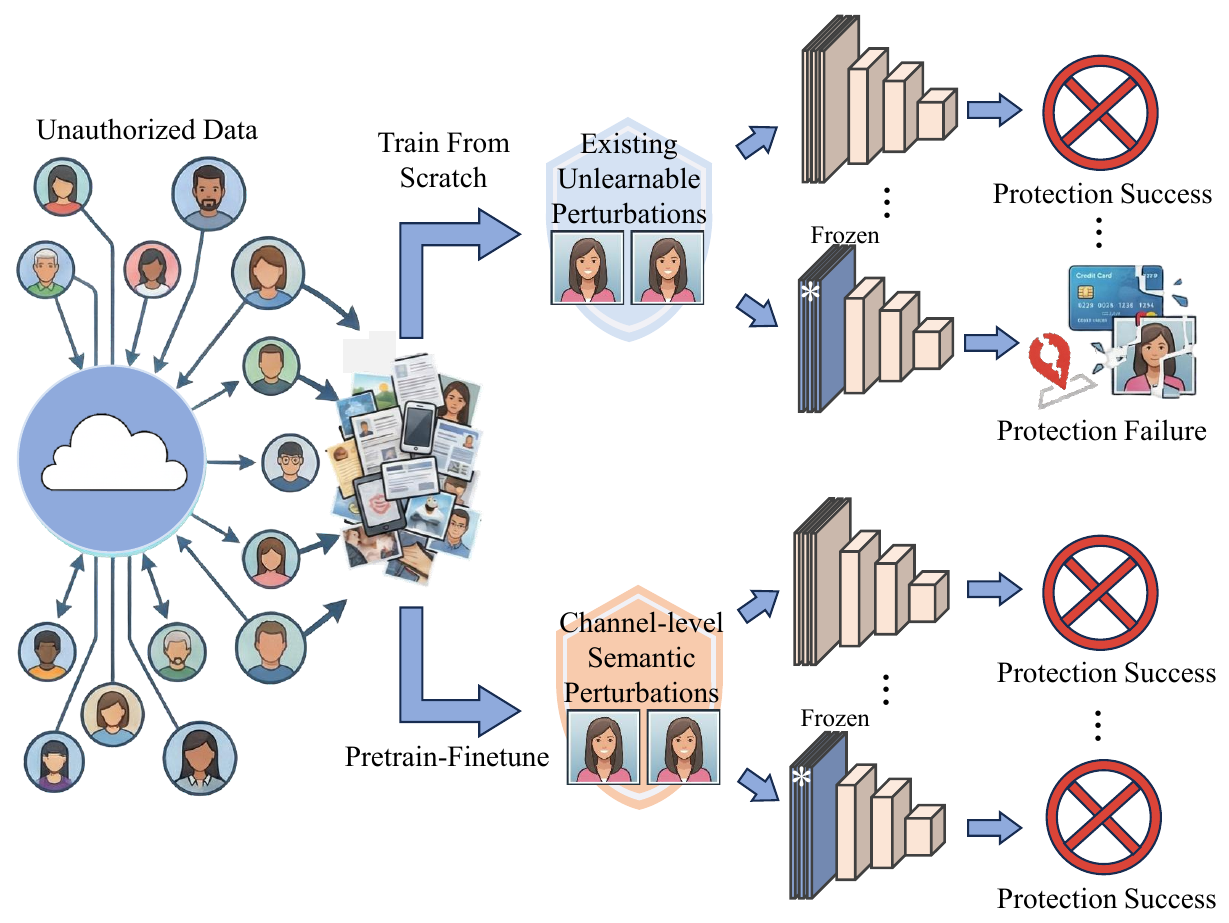}
    \caption{Protection effect of UEs against unauthorized use. 
    To prevent privacy leaks from raw data, unlearnable perturbations are used to inhibit feature learning. While standard perturbations falter under the PF paradigm, the proposed SSC employs channel-level semantic noise to provide robust data protection.}
    \label{fig:unlearnable}
\end{figure}

Early efforts for constructing unlearnable examples originate from the Error-Minimizing Noise paradigm~\cite{EMN}, which suppresses learnable signals by directly reducing the training loss of each example. 
Although the resulting perturbations are visually imperceptible, the unlearnable examples~\cite{EMN, TAP} show vulnerability to adversarial training~\cite{sandoval2023can,yu2022towards
}. 
Meanwhile, data processing before training operations can also reduce the effectiveness of unlearnable perturbations ~\cite{42,44,45}, making it essential to enhance robustness.
To enhance robustness, methods~\cite {REM, wang2021fooling, wen2023adversarial} are proposed, exhibiting increased resistance to adversarial training.
Nevertheless, both Error-Minimizing Noise and Robust Unlearnable Examples suffer from optimization instability. 
To address this issue, Liu et al.~\cite{SEM} further introduced Stable Unlearnable Examples. This method improves the stability of the defense noise by training it against random perturbations.

While recent unlearnable approaches have demonstrated strong effectiveness in obstructing model learning, they mainly focus on the from-scratch training settings. However, their behavior under the widely adopted pretraining–finetuning paradigm~\cite{iofinova2022well,lu2024indiscriminate} remains insufficiently explored. Thus, this work provides a systematic investigation of UEs across diverse training paradigms.
The analyses reveal that loading and freezing pretrained network weights during optimization substantially weakens the unlearnability of existing methods.
In particular, the loading and freezing of the shallow layers of pretrained models significantly affect the protection performance.
Building on this, we further observe that weight-aware perturbations tailored for pretrained weights can mitigate the vulnerability, but these generation methods exhibit limited cross-weight transferability in practical black-box scenarios. Consequently, both normal and weight-aware unlearnable examples remain vulnerable under the PF paradigm.

From the perspective of semantic mismatch, we further provide explanations for these findings. In our view, the pretrained shallow layers extract semantic information, whereas existing unlearnable perturbations are mostly semantically inconsistent with natural images.
This mismatch enables frozen shallow layers to function as reliable semantic filters, ignoring semantic-mismatch information and preserving the structure of natural objects, thus passing more useful information. 
To further validate this hypothesis, we introduce SF-Pretrain, which forces the pretrained network to extract semantic representations, thereby concentrating on natural representation spaces and enhancing semantic filtering capabilities during shallow layer learning. Experimental results demonstrate that loading weights reinforced with semantic priors can more effectively reduce unlearnability.

In this work, to enhance the data protection capabilities of UEs across diverse training paradigms,
we further propose a hierarchical deception strategy, Shallow Semantic Camouflage, preserving the efficacy of unlearnable examples suppressed by pretrained weights and semantic-focused shallow layers.
In contrast to conventional methods that optimize noise solely to induce overfitting, our framework employs a reference model as a semantic guide and applies adversarial constraints to enforce strict semantic alignment in shallow layers to generate channel-level semantic perturbations. This approach compels the optimization process to transfer perturbations from shallow features to higher levels of the representation space.
Within these spaces, the induced perturbations can simulate genuine semantics during propagation. As a result, perturbations bypass shallow filtering and influence deeper semantic processing.
Extensive experiments confirm that our method not only performs robustly under standard transfer settings, but also shows strong robustness under challenging training paradigms such as shallow freezing and SF-Pretrain.
We summarize our main contributions as follows:
\begin{itemize}[leftmargin=*, nosep, topsep=0pt, partopsep=0pt]
    \item We systematically reveal the vulnerability of unlearnable examples under the pretraining–finetuning paradigm. Through a comprehensive analysis across diverse finetuning configurations, we find that loading freezing pretrained shallow layers significantly weakens existing UEs protection.
    \item We systematically explain the reason of UEs failing under the PF paradigm. Through an analysis of feature propagation and frequency-domain behavior, we observe that semantic mismatch with natural image statistics is the primary cause of the resulting degradation in unlearnability.
    \item We propose a hierarchical deception strategy to preserve the unlearnability across diverse training paradigms. By enforcing semantic alignment in shallow layers to generate channel-level semantic perturbations, which can bypass pretrained semantic filters and remain robust.
    \item Extensive experiments on CIFAR-10, CIFAR-100, and Tiny-ImageNet demonstrate that our hierarchical deception strategy consistently outperforms state-of-the-art (SOTA) baselines across diverse training paradigms.
\end{itemize}

\section{Related work}
\subsection{Unlearnable Examples}

Unlearnable examples are a data-centric privacy protection technique. Its core principle is to introduce imperceptible perturbations into training examples, aiming to prevent machine learning models from extracting effective feature representations, thereby protecting data privacy.
Existing research on unlearnable examples can be broadly categorized into three directions according to their optimization objectives: improving protection effectiveness, enhancing perturbation robustness, and increasing task versatility and stealthiness.

The first line of work focuses on improving protection effectiveness. Error-Minimizing noise \cite{EMN} formulates UEs generation as a bi-level optimization problem that induces the model into local optima. To alleviate its computational cost, the Game-Theoretic Unlearnable Example \cite{GUE} reformulates the process as a Stackelberg game with a generator approximating the equilibrium. 
Other approaches enhance effectiveness from different perspectives. Targeted Adversarial Poisoning \cite{TAP} frames unlearnability as a data poisoning problem, while Neural Tangent Generalization Attack \cite{NTGA} exploits Neural Tangent Kernel analysis to induce clean-label generalization failure in black-box settings. The Autoregressive method \cite{AR} further removes reliance on surrogate models by generating perturbations without targeting a specific network.

From the perspective of robustness, previous studies focus on preserving the efficacy of perturbation under data augmentation and defensive training strategies. Robust Error-Minimizing \cite{REM} integrates adversarial training and expectation over transformations into the optimization process to counter augmentations. Stable Error-Minimizing \cite{SEM} introduces consistency regularization into the generation process to maintain consistent unlearnable properties of perturbations across diverse model parameter perturbations and input transformations. Provably Unlearnable Examples \cite{wang2024provably} no longer relies solely on empirical test accuracy; instead, it adopts techniques such as parametric smoothing to derive certified upper bounds on achievable test accuracy, providing theoretical guarantees for the degradation of model performance.

A third research direction investigates versatility and stealthiness across diverse tasks and model settings. Versatile Transferable Generator \cite{VTG} introduces an adversarial domain enhancement strategy that generates perturbations by simulating changes in the data distribution, thereby improving cross-architecture transferability. Multimodal Unlearnable Examples \cite{MEM} introduce collaborative disruption between the visual and linguistic modalities by jointly optimizing image perturbations and textual triggers. Deep Hiding \cite{DH} enhances perceptual stealthiness by embedding predefined semantic patterns into clean examples using invertible neural networks.

\subsection{Existing Defenses against UEs}
Defenses against unlearnable examples aim to restore data utility by reducing the impact of imperceptible perturbations during model training. Existing research can be divided into three primary directions: preprocessing-based defenses, training-phase defenses, and generative purification defenses.

Preprocessing-based defenses leverage the sensitivity of low-level perturbation features to image processing to neutralize UEs through data transformation. Traditional methods utilize lightweight operations such as grayscale conversion, JPEG compression \cite{liu2023image, marcellin2000overview}, and spatial filtering to eliminate the shortcuts introduced by UEs \cite{qin2023learning}. While computationally efficient, these methods are often bypassed by unlearnable examples with complex features, and their effectiveness varies significantly depending on the specific attack mechanism.

Training-phase defenses integrate robustness directly into the model optimization process. Adversarial Training \cite{madry2018towards} is a classic strategy that forces the model to learn robust features. To balance efficiency, UEraser \cite{qin2023learning} applies combinations of augmentations (such as Mixup \cite{MixUp}, CutMix \cite{CutMix}, and CutOut \cite{CutOut}) and loss-maximizing policies to diversify data distributions. Furthermore, specialized objectives such as orthogonal projection \cite{sandoval2024learn} isolate and suppress the influence of unlearnable examples during backpropagation. Despite their robustness, the computational cost of these measures during the training phase is massive, and they may lead to potential performance degradation on clean data \cite{liu2023image}.

Generative purification defenses utilize deep generative architectures to separate or strip away perturbations from clean data. For example, D-VAE \cite{yu2024purify} employs a rate-constrained variational autoencoder for unsupervised perturbation separation, thereby maintaining data integrity while removing harmful signals. Similarly, AVATAR~\cite{dolatabadi2023devil} builds upon the methodology of DiffPure~\cite{nie2022diffusion}, applying diffusion models to counteract intentional perturbations while preserving the essential semantics of training images. However, these approaches depend on complex architectures and require substantial auxiliary data, which increases implementation costs~\cite{10841434}.

Overall, existing defenses balance efficiency, robustness, and deployment cost. Defenses based on preprocessing incur relatively low computational overhead but often struggle to handle complex feature distortions. Training-stage defenses enhance model resilience through optimized learning strategies; however, they are accompanied by substantial computational expenses and performance degradation. Generative purification methods achieve strong restoration quality, yet their reliance on sophisticated architectures and auxiliary data limits their scalability in large-scale deployment scenarios.

Existing generation methods of unlearnable examples primarily focus on the training-from-scratch paradigm, leaving their efficacy under pretraining–finetuning unexplored.
This PF paradigm is both more computationally efficient and more aligned with real-world deployment than training models from the ground up.
Our evaluation reveals that the pretrained models, especially with layer freezing, significantly suppress the unlearnability of existing UEs. Instead, this paper proposes a novel SSC to realize unlearnability under the PF paradigm.
\section{Vulnerability of Unlearnable Examples}

\subsection{Preliminary}
\label{sec:preliminary}
\subsubsection{Unlearnable Examples}
\label{sec:prelim_finetuning}
Unlearnable examples aim to prevent models from using unauthorized examples by reducing the learning performance of the model on perturbed data. Following prior work~\cite{EMN, REM}, we consider a standard $K$-class image classification task. Let $\mathcal{D}_c = \{(x_i, y_i)\}_{i=1}^N$ denote the clean training dataset containing $N$ examples, where $x_i \in \mathcal{X} \subset \mathbb{R}^d$ represents the input image and $y_i \in \mathcal{Y} = \{1, \dots, K\}$ denotes the corresponding label.

To prevent unauthorized model training using the dataset, the protector creates an unlearnable dataset $\mathcal{D}_u$ by adding imperceptible perturbations to the clean images. The unauthorized model denoted by $f(\cdot; \theta): \mathcal{X} \to \mathcal{Y}$, with parameters $\theta$.  The unlearnable dataset is defined as:
\begin{equation}
    \mathcal{D}_u = \{(x_i + \delta_i, y_i)\}_{i=1}^N,
\end{equation}
where $\delta_i \in \mathbb{R}^d$ represents the unlearnable perturbation. To ensure visual stealthiness, the perturbation is limited within a bounded budget, which generally satisfies $\|\delta_i\|_p \le \epsilon$.

Existing methods enable the model to learn perturbations rather than semantic information by adding a shortcut feature form that takes precedence over semantic content.
Specifically, this is achieved by solving a bi-level optimization problem on the surrogate model $f'(\cdot; \theta')$. In this process, the inner optimization shows how a learner updates model parameters using perturbed data, while the outer optimization adjusts the perturbation to hinder learning as much as possible.
\begin{equation}
    \begin{aligned}
        &\min_{\delta} \mathbb{E}_{(x, y) \sim \mathcal{D}_c} [\mathcal{L}(f'(x + \delta; \theta^*), y)],\\
        &\text{s.t.} \quad \theta^* = \arg\min_{\theta'} \mathbb{E}_{(x, y) \sim \mathcal{D}_c} [\mathcal{L}(f'(x + \delta; \theta'), y)],
    \end{aligned}
\end{equation}
where $\mathcal{L}(\cdot)$ denotes the cross-entropy loss function. The perturbation $\delta$ encourages the training loss to decrease more rapidly on shortcut features than on semantically meaningful features during optimization. As a result, the unauthorized model $f$ is more likely to learn non-robust representations, leading to degraded generalization performance on the clean test set.

\subsubsection{\bf PF Paradigm}
\label{sec:prelim_finetuning}

Unlike previous studies that predominantly evaluated unlearnable examples under the training-from-scratch setting, this work investigates the robustness of UEs under the practical pretraining–finetuning paradigm. This setup reflects realistic unauthorized training scenarios, where adversaries typically rely on pretrained weights instead of training models from scratch.

Let $f(\cdot; \theta_{pre})$ denote a model pretrained on a large-scale source dataset. 
Under the PF paradigm, the adversary initializes the unauthorized model parameters $\theta$ with $\theta_{pre}$ to perform the classification task described in Eq.~(2).
To formally analyze the impact of layer-wise transferability, we selected different network components to freeze as distinct finetuning configurations. 
Regardless of the specific freezing strategy employed, the model parameters can be generalized into a binary structure. Specifically, we decompose the model parameters $\theta$ into two functional components:
\begin{equation}
    \theta = \{ \theta_f, \theta_l \},
\end{equation}
where $\theta_f$ (fixed parameters) denotes the subset of weights that are kept frozen during the fine-tuning process, and $\theta_l$ (learnable parameters) represents the remaining weights that are subject to gradient-based optimization. 
This decomposition is particularly important for investigating how the fixed layers of the pretrained model interact with unlearnable perturbations and how the learnable layers optimize during training. 

During the fine-tuning phase, the model loads pre-trained parameters and subsequently updates $\theta_l$ on the unlearnable dataset $\mathcal{D}_u$, expressed as:
\begin{equation}
\theta_l^* = \arg\min_{\theta_l} 
\sum_{(x_i+\delta_i, y_i) \in \mathcal{D}_u} 
\mathcal{L}\big(f(x_i+\delta_i; \theta_f^{pre}, \theta_l), y_i\big).
\end{equation}

\begin{table*}[t]
\centering
\small 
\renewcommand{\arraystretch}{1.3} 
\setlength{\tabcolsep}{10pt} 
\caption{Vulnerability analysis of UEs under the PF paradigm. 
We evaluate the clean test accuracy (\%) of various UEs methods against unauthorized models with different frozen layer configurations. 
The column headers denote the layers frozen during the unauthorized model's training, in which $L_{1,2}$ indicates that Layer 1 and Layer 2 are frozen, while `—' denotes full model training without any frozen layers.
Higher accuracy indicates a greater vulnerability of unlearnable examples.}
\label{tab:vulnerability_trend}
\begin{tabular}{c|ccccccccccc}
\toprule
\multirow{2}{*}{\bf Method} & \multicolumn{10}{c}{\bf Unauthorized Model Frozen Layers} \\ \cmidrule(l){2-12} 
 & $\mathcal{L}_{3,4}$ & $\mathcal{L}_{2,4}$ & $\mathcal{L}_{2,3}$ & $\mathcal{L}_{1,4}$ & $\mathcal{L}_{1,3}$ & \boldmath$\mathcal{L}_{1,2}$ & $\mathcal{L}_{4}$ & $\mathcal{L}_{2}$ & $\mathcal{L}_{3}$ & $\mathcal{L}_{1}$ & — \\ \midrule
 \rowcolor{gray!20} 
{\bf Clean} & 93.42 & 95.18 & 95.06 & 92.73 & 92.29 & 90.54 & 94.11 & 95.37 & 93.68 & 95.21 & 93.40 \\ 
{\bf EMN} & 43.15 & 46.82 & 54.39 & 46.51 & 52.07 & 64.93 & 42.26 & 49.74 & 47.12 & 55.48 & 18.47\\
 \rowcolor{gray!20} 
{\bf NTGA} & 14.57 & 27.24 & 39.81 & 53.06 & 59.43 & 67.12 & 26.39 & 24.55 & 21.08 & 29.34 & 11.83\\
{\bf REM} & 24.83 & 30.12 & 31.55 & 31.92 & 31.04 & 45.76 & 29.87 & 29.91 & 29.74 & 29.38 & 17.25\\
 \rowcolor{gray!20} 
{\bf AR} & 12.41 & 14.63 & 17.29 & 27.84 & 38.15 & 79.27 & 14.33 & 14.59 & 18.02 & 33.61 & 13.91\\
{\bf SHR} & 13.96 & 21.48 & 32.05 & 28.37 & 28.61 & 43.19 & 14.72 & 22.14 & 12.85 & 30.52 & 12.06\\ \bottomrule
\end{tabular}
\end{table*}

To thoroughly evaluate the robustness of Unlearnable Examples under the PF paradigm, we conduct quantitative experiments on the CIFAR-10 dataset using the ResNet-18 architecture. 
In our experimental configuration, $L_i$ denotes the $i$-th residual block of the model. For example, $L_{3,4}$ indicates that the first two blocks remain trainable, while the third and fourth blocks are fixed with pre-trained weights. We measure the clean test accuracy of the unauthorized model to evaluate defense effectiveness: a higher accuracy indicates a greater vulnerability of the UEs.
The quantitative results are summarized in Table~\ref{tab:vulnerability_trend}. Our key observations are as follows:
\begin{enumerate}
    \item Unlearnable examples demonstrate significant effectiveness under the training-from-scratch paradigm. For instance, training on EMN  yields a clean test accuracy of 18.47\%, indicating that model generalization is successfully prevented in the absence of pretrain parameters.
    
    \item All evaluated UEs methods exhibit significant vulnerabilities under the PF paradigm.
    The protection fails substantially when foundational components are included in the fixed parameters $\theta_f$. This widespread failure across all evaluated methods suggests that pretrained features can effectively bypass the protection intended by UEs.
    
    \item Freezing shallow layers is more detrimental to unlearnable examples protection than freezing deep layers. For the AR method, the $L_{1,2}$ configuration reaches 79.27\% accuracy while the $L_{3,4}$ configuration only reaches 12.41\% under. This phenomenon can also be observed in other defense methods, confirming that shallow-layer features play a dominant role in neutralizing the defense.
\end{enumerate}
\begin{figure}[t]
    \centering
    \includegraphics[width=0.9\linewidth]{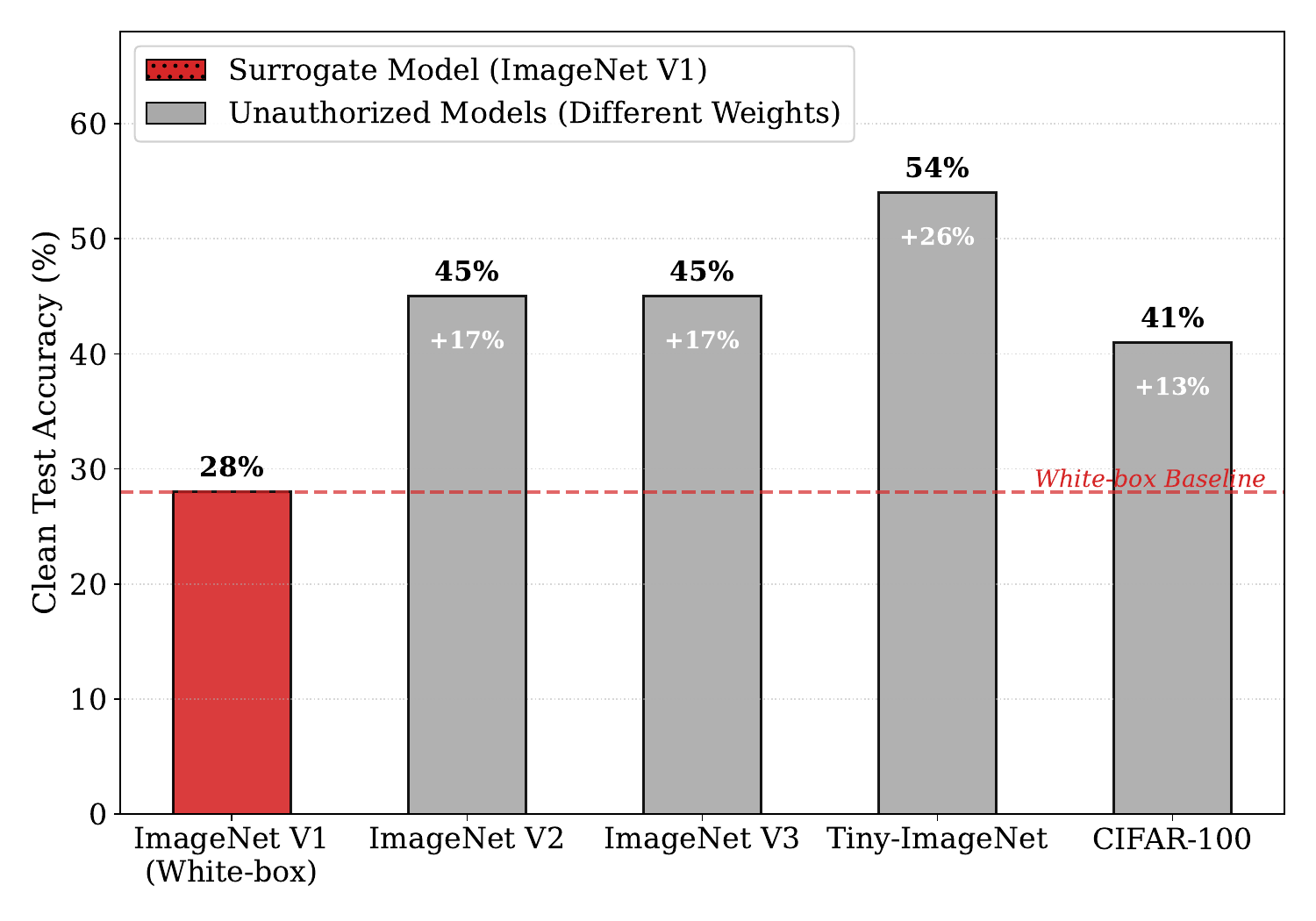}
    \caption{Evaluation of cross-weight robustness in black-box settings. 
Evaluating the transferability of weight-aware unlearnable examples (optimized on ImageNet V1 weights) against models initialized with different weight configurations.  
The dashed line marks the theoretical lower bound of the attack.}
    \label{fig:cross_weight}
\end{figure}
\subsection{Vulnerabilities of Weight-Aware Unlearnable Generation}
\label{subsec:adaptive_generation}
Based on the observation that frozen shallow layers significantly attenuate unlearnability, a natural question arises: Can data protectors restore effective protection under the PF paradigm by incorporating pretrained models into the construction of unlearnable examples? To answer this, we conduct a weight-aware unlearnable generation experiment, investigating the extent to which this method can counteract the observed protection failure.
To systematically evaluate the effectiveness of these adaptive attacks, we generated a series of unlearnable examples by unfreezing different combinations of residual blocks
, resulting in 10 distinct generation strategies, ranging from shallow-focused optimization to deep-focused optimization and multi-scale hybrid configurations.

The performance of these strategies under white-box settings, where the surrogate and unauthorized model share identical weights, is presented in Table~\ref{tab:adaptive_attack}. Furthermore, the black-box experimental results, which evaluate the transferability of perturbations optimized on ImageNet V1 against different weight initializations, are shown in Figure~\ref{fig:cross_weight}. Based on a comprehensive comparison and analysis, the primary experimental observations are presented as follows:
\begin{table*}[t]
\centering
\small
\renewcommand{\arraystretch}{1.3} 
\setlength{\tabcolsep}{9pt} 
\caption{Vulnerability analysis of weight-aware UEs.
We evaluated the clean test accuracy (\%) of UEs generated by 10 freezing strategies under 4 different weight initialization configurations.
The column headers denote the trainable layers of the surrogate model during generation. The row headers indicate the trainable layers of the unauthorized model during training.}
\label{tab:adaptive_attack}
\begin{tabular}{l|cccccccccc}
\toprule
\multirow{2}{*}{\bf Unauthorized Model} 
& \multicolumn{10}{c}{\bf Surrogate Model Trainable Layers} \\ 
 & $\mathcal{L}_{1,2}$ & $\mathcal{L}_{1,3}$ & $\mathcal{L}_{1,4}$ & $\mathcal{L}_{2,3}$ & $\mathcal{L}_{2,4}$ & $\mathcal{L}_{3,4}$ & $\mathcal{L}_{1,2,3}$ & $\mathcal{L}_{1,3,4}$ & $\mathcal{L}_{1,2,4}$ & $\mathcal{L}_{2,3,4}$ \\ \midrule
 \rowcolor{gray!20} 
$\mathcal{L}_{1,2}$ + FC & 53.24 & 48.17 & 33.59 & 41.02 & 13.46 & 13.81 & 23.94 & 34.07 & 12.35 & \textbf{10.28} \\
$\mathcal{L}_{2,3}$ + FC & 64.82 & 70.33 & 48.61 & 44.15 & 25.49 & 30.72 & 33.16 & 43.88 & 34.21 & \textbf{11.54} \\
\rowcolor{gray!20} 
$\mathcal{L}_{3,4}$ + FC & 80.11 & 79.56 & 72.48 & 62.37 & 68.92 & 48.05 & 66.23 & 53.64 & 69.47 & \textbf{28.19} \\
$\mathcal{L}_{2,3,4}$ + FC & 68.73 & 73.41 & 52.06 & 47.85 & 27.29 & 32.64 & 35.50 & 44.12 & 41.93 & \textbf{13.76} \\ \bottomrule
\end{tabular}%
\end{table*}
\begin{enumerate}
\item Models trained on weight-aware UEs consistently achieve lower clean test accuracy than those trained on standard UEs under the PF setting. This suggests that incorporating pretrained weights during generation can partially alleviate the degradation of unlearnability.
\item In the white-box setting, freezing conv1 and layer1 during generation yields the strongest protection performance across all configurations. This indicates that constraining perturbation optimization in shallow layers biases the generation process toward perturbations more robust to layer-freezing. As a result, these perturbations become more transferable under adaptive attacks.
\item When the weights used during generation precisely match the unauthorized model's initialization (ImageNet V1), the unlearnable examples preserve their unlearnability, reducing accuracy to 28\%. This confirms that the perturbation effectively exploits the specific numerical distribution of the weights for which it was optimized.
\item In the black-box setting, when the attack is transferred to models initialized with different weights, such as those pretrained on ImageNetV2/V3 and CIFAR-100, its effectiveness declines markedly, with test accuracy increasing to 45\% or higher. This finding indicates that the weight-aware generation strategy overfits to particular parameter values rather than learning weight-agnostic features.
\end{enumerate}

\textit
{In summary, the results demonstrate that while weight-aware unlearnable generation can maintain unlearnable performance by leveraging specific pretrained parameters, these perturbations lack generalization under black-box settings. 
Given that white-box scenarios do not exist in reality, these findings indicate that enhancing the robustness of unlearnable examples under the PF paradigm cannot depend exclusively on leveraging specific pretrained weights.}

\section{Failure Mechanisms of Unlearnable Examples}
\label{sec:diagnosis}

\label{sec:feature_silence}
In this section, we explore the reasons why unlearnable examples fail under the pretraining-finetuning paradigm when shallow layers remain frozen.
First, we evaluated the inconsistency of layer-wise features and found that features exhibited unusually high consistency in pre-trained models with frozen shallow layers; this indicates that perturbations fail to dominate feature representations, unlike the patterns observed in models trained from scratch. Then, we investigated the perturbation transmission dynamics and discovered that semantic mismatch significantly suppresses perturbation energy within these shallow layers. Additionally, further analysis pinpointed this mismatch to differences in frequency-domain distributions: pre-trained shallow layers function as semantic filters that primarily respond to low-frequency natural image semantics while automatically filtering out mid-to-high frequency perturbation signals. Finally, we validated this analysis through SF-Pretrain experiments.

\subsection{Noise Transmission Analysis under PF}
To quantitatively describe the impact of unlearnable perturbations at different network depths, we first examine their effects at the representation level via feature-space consistency.
Let $f_l(\cdot)$ denote the feature map output by the $l$-th layer of the model. For a clean example $x$ and its unlearnable counterpart $x' = x + \delta$, we measure the layer-wise cosine similarity:
\begin{equation}
    \mathcal{S}_l(x, \delta) = 
    \frac{\langle \mathrm{vec}(f_l(x)), \mathrm{vec}(f_l(x+\delta)) \rangle}
    {\|\mathrm{vec}(f_l(x))\|_2 \cdot \|\mathrm{vec}(f_l(x+\delta))\|_2},
\end{equation}
where $\mathrm{vec}(\cdot)$ flattens the feature map into a vector.

\begin{figure}[t]
    \centering
    \includegraphics[width=1\linewidth]{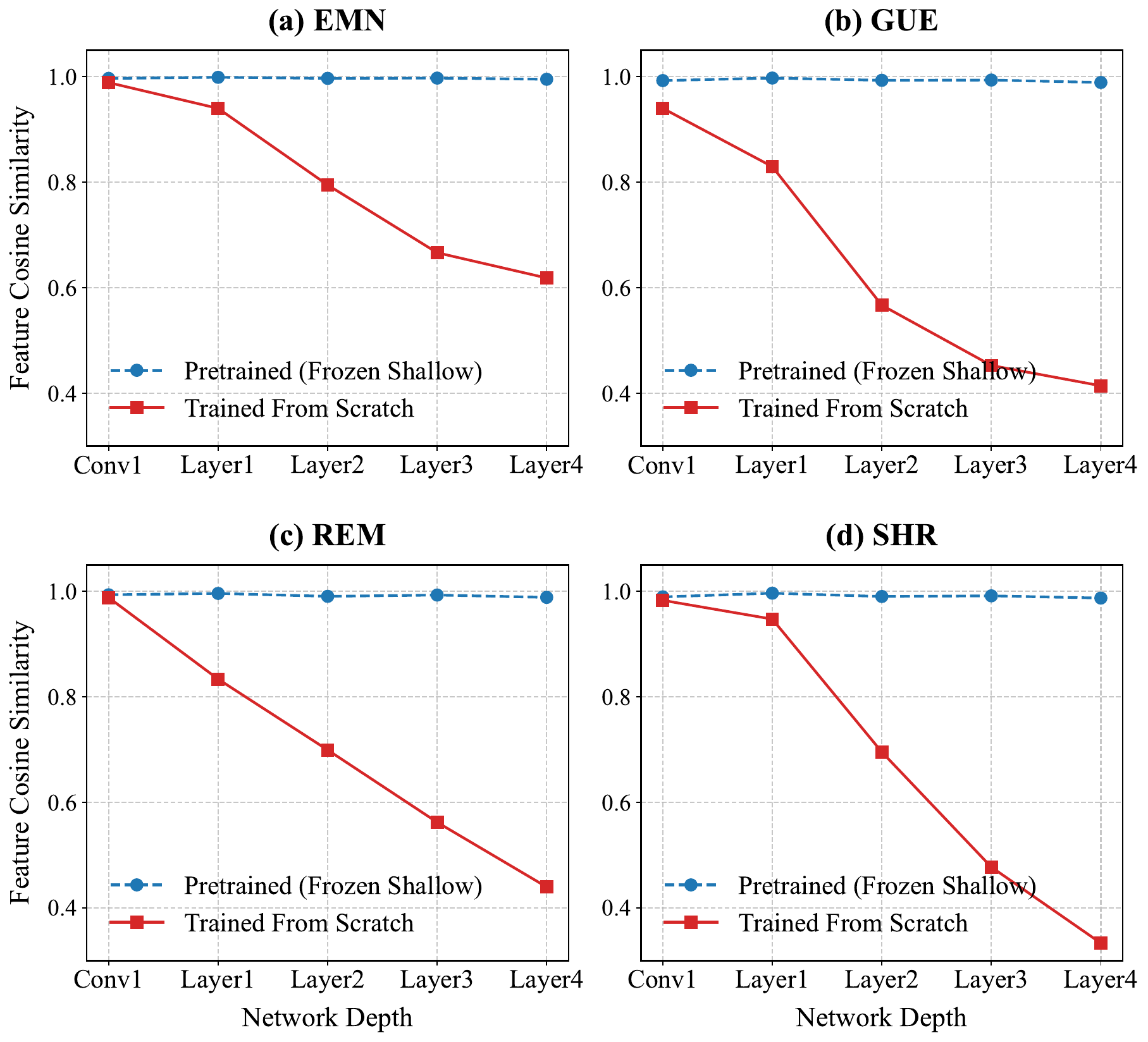} 
    \caption{Layer-wise feature consistency. Cosine similarity $\mathcal{S}_l$ between clean and unlearnable feature representations across network depth. While perturbations progressively decouple features in models trained from scratch, pretrained models with frozen shallow layers  maintain consistently high similarity.}
    \label{fig:feature_consistency}
\end{figure}

As shown by the red line in Fig.~\ref{fig:feature_consistency}, the semantic consistency $\mathcal{S}_l$ exhibits a pronounced declining trend as the network depth increases when trained from scratch.
In contrast, under the PF paradigm with frozen shallow layers as shown
by the blue line, we observe a strikingly different phenomenon: despite the injection of perturbations capable of deceiving models trained from scratch, feature representations remain highly consistent throughout the network ($\mathcal{S}_l > 0.98$ in stage~1--2 of ResNet).

However, consistency at the feature level alone does not clarify whether the perturbation was successfully injected and subsequently ignored, or whether it failed to propagate through the network.
To directly examine the propagation behavior of unlearnable perturbations, we analyze the dynamics of signal transmission across network layers.

To better characterize this propagation, we introduce the Perturbation Transfer Rate (PTR), a metric that quantifies the sensitivity of feature representations to injected noise, defined as follows:
\begin{equation}
    \mathrm{PTR} = \frac{\|\Phi(x+\delta) - \Phi(x)\|_2}{\|\Phi(x)\|_2}.
\end{equation}

\begin{figure}[t]
    \centering
    \includegraphics[width=1\linewidth]{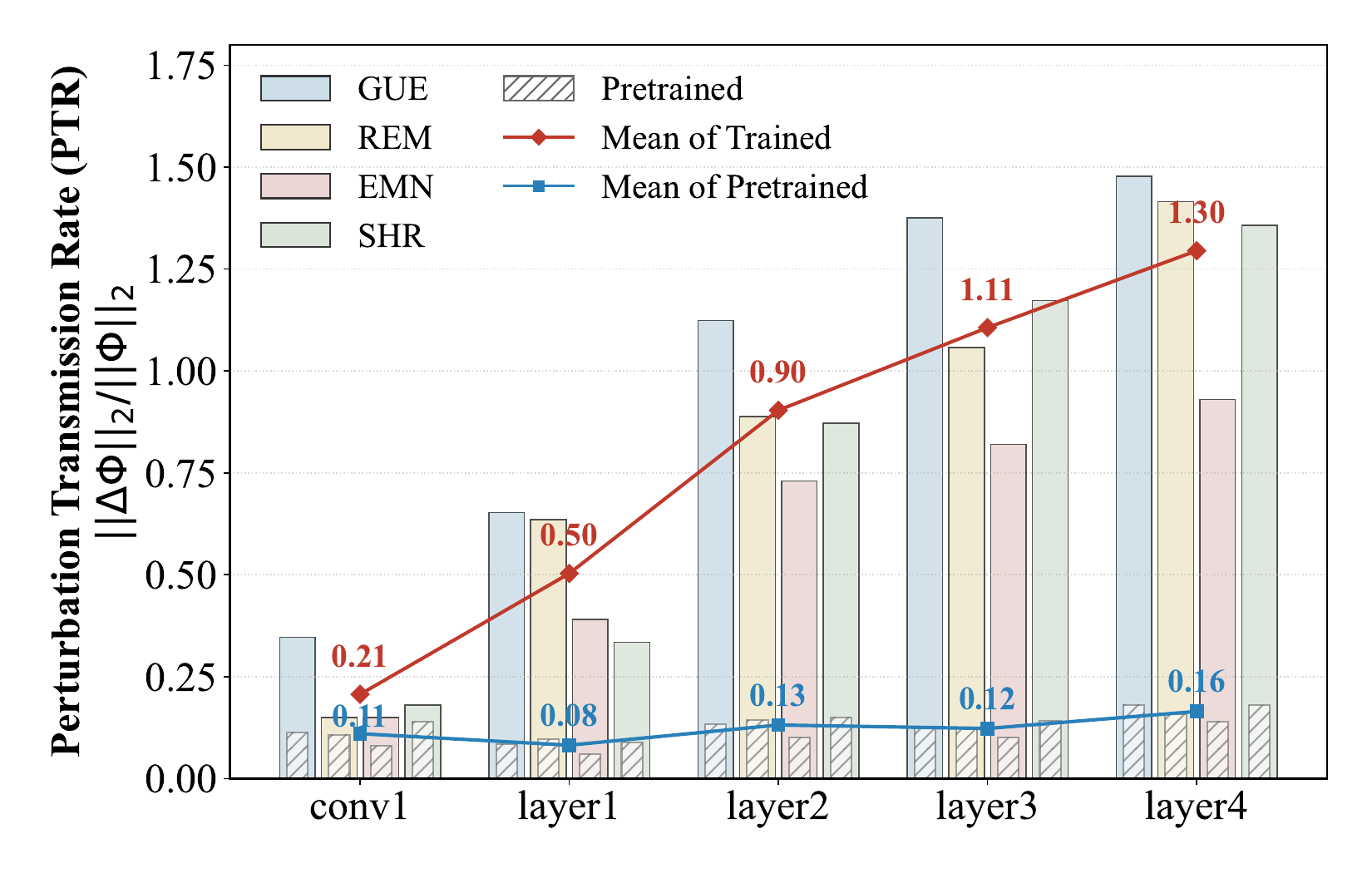}
    \caption{Perturbation transmission analysis. Solid bars represent UEs trained from scratch, showing a progressive amplification of perturbation energy in deeper layers. In contrast, hatched inner bars denote pretrained models, where the energy is attenuated. The line indicates the mean PTR trend of models.}
    \label{fig:filtering_mechanism_final}
\end{figure}

Fig.~\ref{fig:filtering_mechanism_final} illustrates the contrasting behaviors of a model trained from scratch on unlearnable examples and a pretrained model with frozen shallow layers.
Unlearnable noise demonstrates distinct transmission patterns under these two training paradigms.
The model trained from scratch exhibits a steadily increasing PTR, confirming that perturbations are amplified during training and subsequently become the primary shortcut features.
In contrast, the pretrained model demonstrates significant suppression of perturbation energy. This attenuation emerges in the initial shallow layer and persists throughout the network. These findings suggest that unlearnable perturbations generated by different methods do not propagate beyond frozen, pretrained, shallow layers, thereby preventing their influence on downstream learnable parameters.
Consequently, the feature inconsistency observed in 
Fig.~\ref{fig:feature_consistency}
results directly from perturbation transmission failure, rather than from insensitivity of the pretrained model to the features. 

\subsection{Analysis of Semantic Filtering}
\label{sec:semantic_filtering}

\begin{figure}[t]
    \centering
    \includegraphics[width=1\linewidth]{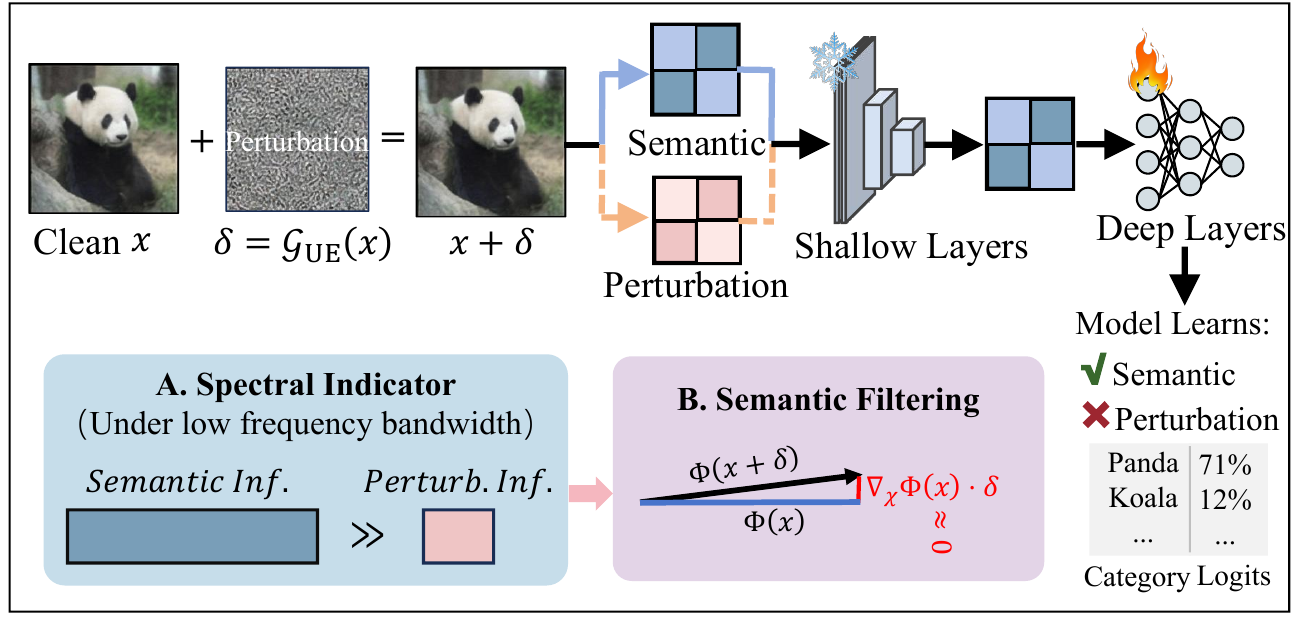} 
    \caption{Illustration of the Semantic Filtering Effect. This figure demonstrates that the inherent semantic information in the input is naturally stronger than the perturbations. The frozen shallow layers of the model act as a filter, further suppressing noise and amplifying the dominance of semantic features.}
    \label{fig:semantic_filtering}
\end{figure}
The preceding analysis demonstrates that frozen, pretrained shallow layers inhibit the propagation of unlearnable perturbations. However, the precise physical origin of this semantic filtering and why these perturbations fail to bypass the shallow layers remains unclear.
To address this, we perform a spectral analysis, which revealed that the pretrained shallow layers act as semantic filters due to the statistical mismatch between the perturbations and the semantics of natural images.

\begin{figure*}[t]
    \centering
    \begin{minipage}{0.66\linewidth}
        \centering
        \includegraphics[width=1.02\linewidth]{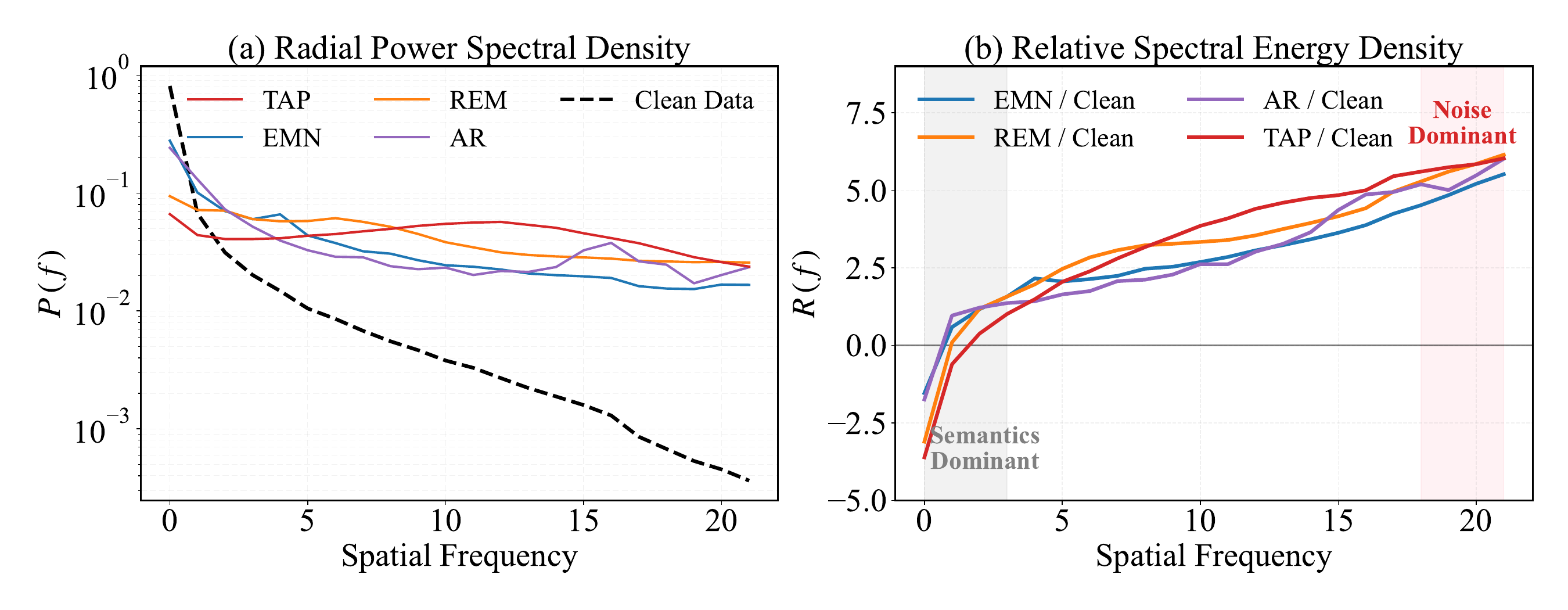}
        \captionof{figure}{Spectral analysis of UEs. (a) Radial PSD of various examples, where the power density $P(f)$ is defined in Eq.~(8). (b) Relative spectral energy density $R(f)$ between perturbations and clean examples, defined in Eq.~(9) as the $\log_2$ ratio.}
        \label{fig:relative_spectral_density}
    \end{minipage}%
    \hfill
    \begin{minipage}{0.32\linewidth}
        \centering
        \includegraphics[width=0.88\linewidth]{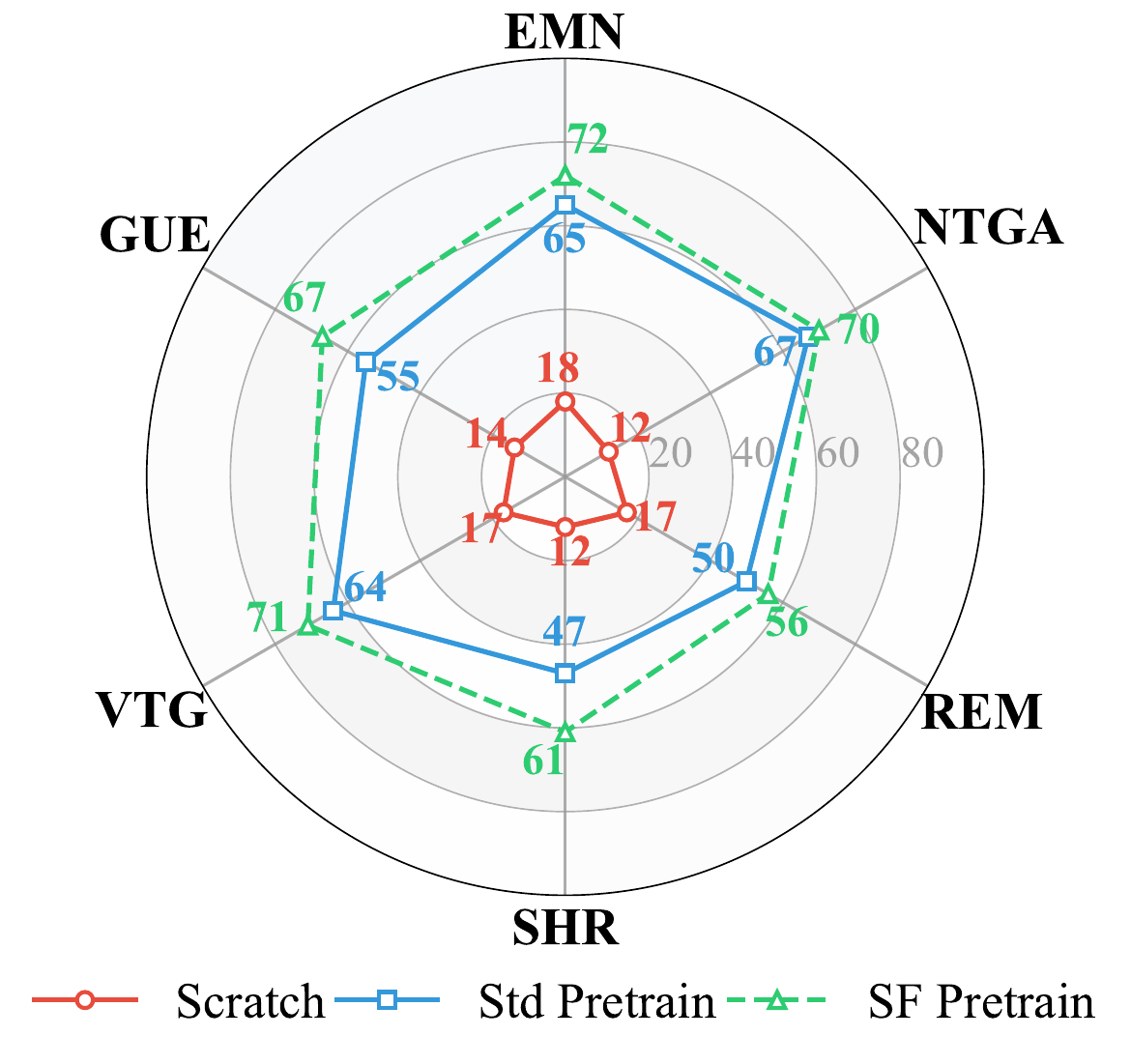}
        \captionof{figure}{Robustness comparison of UEs under diverse training paradigms.
        UEs exhibit vulnerability under SF-Pretrain.}
        \label{fig:verification_experiment}
    \end{minipage}
\end{figure*}

The response of shallow layers $\Phi(\cdot; \theta_{\text{s}})$ can be approximated via a first-order Taylor expansion:
\begin{equation}
    \Phi(x + \delta) \approx \Phi(x) + \nabla_x \Phi(x) \cdot \delta,
\end{equation}
Under this approximation, the resulting feature discrepancy induced by perturbation $\delta$ is governed by the projection term $\|\nabla_x \Phi(x) \cdot \delta\|_2$, reflecting its preservation through shallow layers. A spectral mismatch between the perturbation energy and the frequency components of natural images leads to discrepancies in shallow-layer gradient interactions, which in turn lead to semantic filtering.
To quantitatively characterize this mismatch, we employ Power Spectral Density (PSD) analysis in the frequency domain.
We compute the radially averaged PSD, denoted as $P(f)$, by averaging the 2D power spectrum $| \mathcal{F}(z)(u, v) |^2$ over the azimuthal angle $\phi$ for a given radial frequency $f = \sqrt{u^2 + v^2}$:\begin{equation}P(f) = \frac{1}{2\pi} \int_{0}^{2\pi} | \mathcal{F}(z)(f, \phi) |^2 d\phi.\end{equation}
Here, $z$ denotes an input (either a clean example $x$ or a perturbation $\delta$).
This function $P(f)$ measures the energy distribution across spatial frequencies $f$.
As illustrated in Fig.~\ref{fig:relative_spectral_density}(a), the radially averaged PSD $P(f)$ reveals a fundamental divergence in energy distribution between natural images and perturbations. 
For a more accurate comparison, we define the relative spectral density between the perturbation and the natural image as:
\begin{equation}
R(f) = \log_2 \left( \frac{P_{\delta}(f)}{P_{x}(f)} \right),
\end{equation}
where $P_{\delta}(f)$ and $P_{x}(f)$ denote the radially averaged PSD of the perturbation $\delta$ and the clean image $x$, respectively. 

As shown in Fig.~\ref{fig:relative_spectral_density}(b), clean images exhibit a characteristic power-law decay, concentrating energy in low-frequency components associated with semantic structure. In contrast, unlearnable perturbations show reduced energy in these low-frequency bands and a relative dominance in the mid-to-high frequency range. Because shallow kernels $\Phi$ are pretrained to capture natural image regularities, their input gradients $\nabla_x \Phi(x)$ predominantly reside in a subspace aligned with these low-frequency semantic structures. Consequently, as the perturbation $\delta$ deviates from these spectral patterns, its interaction with the shallow-layer gradients weakens, leading to a minimal projection:
\begin{equation}
\|\nabla_x \Phi(x) \cdot \delta\|_2 \ll \|\Phi(x)\|_2.
\end{equation}


These results confirm that frozen pretrained shallow layers function as semantic filters. By suppressing spectral components inconsistent with natural image statistics, they prevent perturbations from propagating to deeper learnable parameters. This ensures that updates to $\theta_{\text{deep}}$ are driven by authentic semantics, disrupting the shortcut pathways exploited by UEs.

\subsection{Analysis of Shallow Semantic Enhancement}
\label{sec:semantic_verification}

\begin{figure*}[t]
    \centering
    \includegraphics[width=\linewidth]{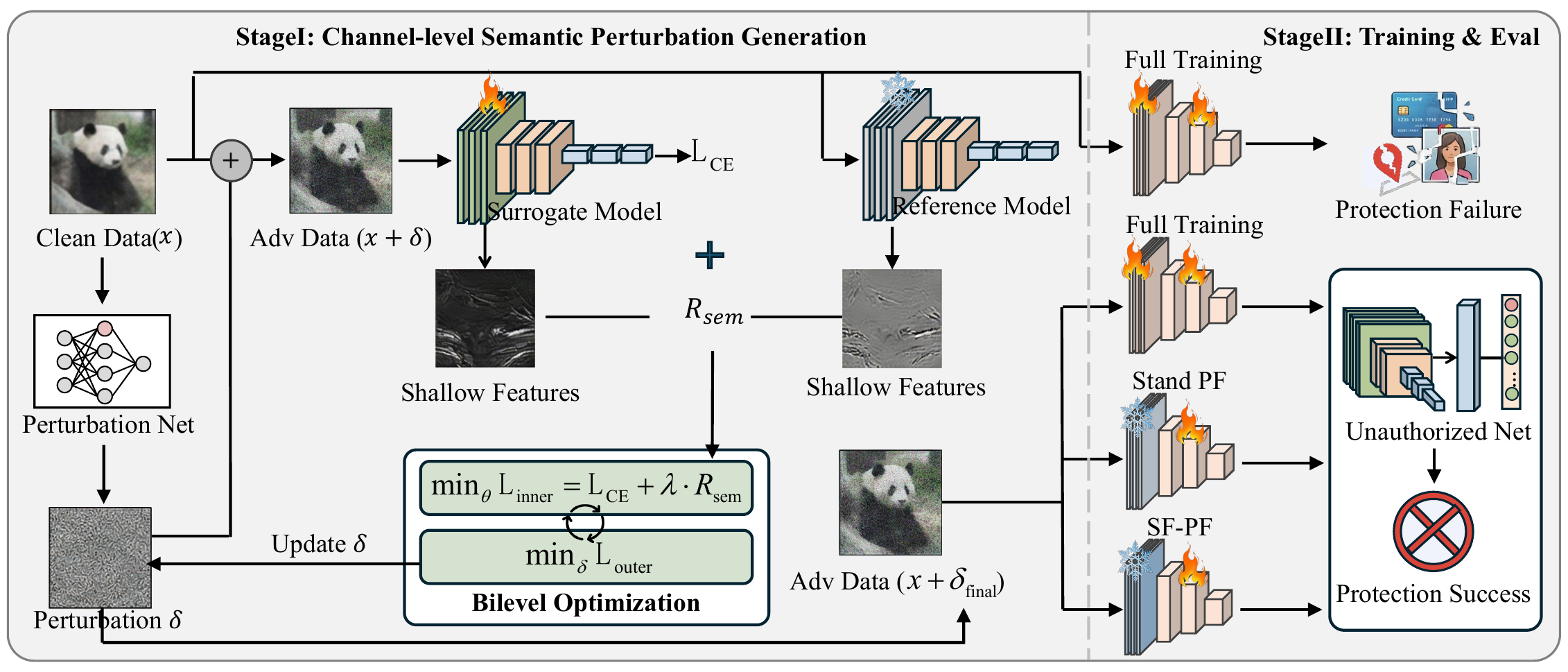} 
    \caption{Overview of the proposed Hierarchical Deception framework via Shallow Semantic Camouflage. 
    After creating noise $\delta$, we introduce a reference anchor to enforce semantic alignment. Noise is anchored to the semantic features of a reference model. Through a Bilevel Optimization Loop, the attack simultaneously updates the surrogate model's weights and the perturbation while maintaining Semantic Alignment ($R_{sem}$).}
    \label{fig:method_framework}
\end{figure*}
To further validate the semantic filtering explanation, this part introduces SF-Pretrain, which enhances semantic awareness in pretrained shallow layers. Specifically, SF-Pretrain introduces constraints in the shallow layers of the model, prompting the shallow network to extract more discriminative features.
Let $\{\Phi_k(x; \theta_{s}^{(k)})\}_{k=1}^{K}$ denote the shallow feature extractors at $K$ distinct early stages.
We attach a lightweight auxiliary classifier $h_{aux}^{(k)}(\cdot)$ to each shallow exit, the enhancement objective is formulated as:
\begin{equation}
\begin{aligned}
    \mathcal{L}_{SF} = &\mathcal{L}_{task}(f(x;\theta), y) + \\
    &\lambda \sum_{k=1}^{K} 
    \mathcal{L}_{CE}^{(k)}\left(h_{aux}^{(k)}(\Phi_k(x; \theta_{s}^{(k)})), y\right),
\label{eq:probe_loss}
\end{aligned}
\end{equation}
where $\mathcal{L}_{task}$ is the standard classification loss of the final output, and $\lambda$ is a penalty coefficient to prioritize shallow semantic alignment.
From an information-theoretic perspective, minimizing the joint semantic loss $\sum \mathcal{L}_{CE}^{(k)}$ can be interpreted as encouraging higher mutual dependence between the shallow features at multiple scales and the semantic labels $y$, which is closely related to maximizing a variational lower bound of the Mutual Information (MI):
\begin{equation}
\max_{\theta_{s}} \sum_{k=1}^{K} I(\Phi_k(x; \theta_{s}^{(k)}); y).
\end{equation}
By explicitly promoting semantic alignment across shallow layers, SF-Pretrain effectively enhances the semantic selectivity of shallow feature extractors. 
SF-Pretrain compels $\theta_{s}$ to eliminate information weakly correlated with the semantic class $y$ at the outset of the network forward pass.
We initialize the unauthorized model using the SF-Pretrain weights and evaluate its robustness against SOTA unlearnable methods.

As visualized in the radar chart in Fig.~\ref{fig:verification_experiment}, the results provide strong support for our hypothesis. Compared to standard pretraining and training from scratch, the red line and the green line, models initialized with SF-Pretrain (green) exhibit a reduction in unlearnability and achieve substantially higher clean test accuracy across all evaluated attack benchmarks.

These findings indicate that substantial enhancement of shallow semantics leads to more thorough filtering of unlearnable perturbations.
These results conclusively demonstrate that the failure of existing methods is closely associated with semantic filtering, thereby motivating the development of the proposed Hierarchical Deception strategy to bypass this filter.

\section{SSC for Hierarchical Deception}
\label{sec:method}
Based on the preceding analysis, the vulnerability of existing unlearnable examples stems from a semantic mismatch between perturbation and the natural images. We propose a hierarchical deception framework termed SSC. 
Our approach aims to achieve protection by aligning channel-level semantic perturbations with the shallow semantic manifold of natural images, enabling them to bypass filters and influence the learning process of deeper layers across diverse training paradigms.

The overview of SSC is shown in Fig.~\ref{fig:method_framework}. SSC begins with generation and semantic anchoring, where a trainable generator produces a perturbation $\delta$ to construct an adversarial example. The example is then processed through both a surrogate model and a frozen reference model to extract shallow features, which are used to calculate a semantic alignment loss ($R_{sem}$). Within the Bilevel Optimization Loop, the framework iteratively updates the surrogate model's weights and the perturbation $\delta$ to ensure the noise remains semantically consistent while achieving the poisoning goal. During the evaluation stage, the optimized perturbations are introduced to the clean dataset, inducing learning failure in the unauthorized model across diverse training paradigms.

The proposed SSC framework employs a dual-objective bilevel optimization to induce hierarchical deception.
Given a surrogate model $f'(\cdot; \theta')$ with parameters $\theta' = \{\theta'_s, \theta'_d\}$, where $\theta'_s$ denotes the shallow parameters and $\theta'_d$ represents the deep parameters.
We introduce a dual-objective inner loop that minimizes classification error and feature discrepancy on perturbed data.
The optimization problem is defined as:
\begin{equation}\label{eq13}
    \min_{\delta} \mathcal{L}_{outer}(\theta^*, \delta) \quad \text{s.t.} \quad \theta^* = \arg\min_{\theta} \mathcal{L}_{inner}(\theta', \delta),
\end{equation}
The perturbation $\delta$ is constrained within the $l\infty$-norm ball $\Delta = \{ \delta \mid \|\delta\|_\infty \le \epsilon \}$, where $\text{Proj}_{\epsilon}$ denotes the projection operator that clips the perturbation to the $\epsilon$-boundary.

The outer objective $\mathcal{L}_{outer}$ targets unlearnability by minimizing the classification loss on the perturbed examples:
\begin{equation}\label{eq14}
\mathcal{L}_{\mathrm{outer}}(\theta^*, \delta) = \sum_{(x_i, y_i) \in \mathcal{D}_c} \mathcal{L}_{\mathrm{CE}}\big(f'(x_i + \delta_i; \theta^*), y_i\big).
\end{equation}

By minimizing $\mathcal{L}_{outer}$, the framework encourages the surrogate model to rely on non-robust features, thereby preventing the model from learning the underlying clean semantic patterns.
Concurrently, we introduce a semantic alignment term to the standard objective in the inner loop to ensure that these perturbations can penetrate pretrained filters. 
The modified inner loss $\mathcal{L}_{inner}$ is formulated as:
\begin{equation}\label{eq15}
\begin{aligned}
\!\!\!\!\!\mathcal\quad{L}_{\mathrm{inner}}(\theta', \delta) = \sum_{(x_i, y_i) \in \mathcal{D}_c} \Big[ &\mathcal{L}_{\mathrm{CE}}\big(f'(x_i + \delta_i; \theta'), y_i\big)\\
&\!\!\!\!\!\!\!\!\!\ + \lambda \cdot \mathcal{R}_{sem}\big( g_{\theta'_s}(x_i + \delta_i), g_{\mathrm{ref}}(x_i) \big) \Big],
\end{aligned}
\end{equation}

where $\mathcal{L}_{\mathrm{CE}}$ denotes the cross-entropy loss. For notational simplicity, we use $g_{\theta_s}(\cdot)$ to denote the aggregate shallow feature extractor corresponding to the set of shallow layers defined in Section~\ref{sec:semantic_verification}. 
The term $\mathcal{R}_{sem}$ quantifies the discrepancy between the surrogate model's shallow features and the semantic anchor established by a frozen reference model $g_{\mathrm{ref}}$. In practical applications, $\mathcal{R}_{sem}$ is computed using Gram matrices of feature activations. These matrices capture channel-wise correlations and texture-level statistics enforced by the reference model within shallow representations. By constraining these Gram-based statistics, the optimization ensures that generated perturbations remain consistent with the semantic priors imposed by the pretrained shallow filters. This dual-objective approach avoids the suppression of semantic-mismatch perturbations during forward propagation.

By introducing channel-layer semantic perturbation, SSC constrains shallow representations toward the natural image manifold, effectively shifting interference to deeper representations. This misleads the model's learning process, preserving the unlearnability across diverse training paradigms.

\begin{algorithm}[t]
\caption{Shallow Semantic Camouflage (SSC)}
\label{alg:SSC_method}
\begin{algorithmic}[1]
\STATE \textbf{Input:} Clean dataset $\mathcal{D}_c$, reference extractor $g_{ref}$, surrogate model $f'(\cdot; \theta')$, budget $\epsilon$, penalty $\lambda$, inner steps $K$, epochs $T$, batch size $B$, learning rates $\alpha, \eta$.
\STATE \textbf{Output:} Unlearnable Dataset $\mathcal{D}_u$
\STATE \textit{\textbf{// Phase I: Initialization}}
\STATE $\delta_i \gets \mathbf{0}$ for all $(x_i, y_i) \in \mathcal{D}_c$; 
\STATE Initialize $\theta' = \{\theta'_s, \theta'_d\}$;
\STATE \textit{\textbf{// Phase II: Optimization Loop}}
\FOR{epoch $t = 1$ to $T$}
    \FOR{minibatch $\{(x_i, y_i)\}_{i=1}^B \subset \mathcal{D}_c$}
        \STATE $\theta'^{(t)} \gets \theta'$; \quad \textit{// Save current state for parameter reset}
        
        \STATE \textit{// Inner Loop: Semantic-Constrained Training}
        \FOR{$k = 1$ to $K$}
            \STATE $z_{adv} \gets g_{\theta'_s}(x_i + \delta_i)$; \quad $z_{ref} \gets g_{ref}(x_i)$;
            \STATE $\mathcal{R}_{sem} \gets \frac{1}{B} \sum_{i=1}^B \| \text{Gram}(z_{adv}) - \text{Gram}(z_{ref}) \|_F^2$;
            \STATE $\mathcal{L}_{inner} \gets \frac{1}{B} \sum \mathcal{L}_{CE}(f'(x_i + \delta_i; \theta'), y_i) + \lambda \mathcal{R}_{sem}$;
            \STATE $\theta' \gets \theta' - \alpha \nabla_{\theta'} \mathcal{L}_{inner}$;
        \ENDFOR
        \STATE $\theta^* \gets \theta'$;
        \STATE \textit{// Outer Loop: Perturbation Update (Poisoning)}
        \STATE $\mathcal{L}_{outer} \gets \frac{1}{B} \sum_{i=1}^B \mathcal{L}_{CE}(f'(x_i + \delta_i; \theta^*), y_i)$;
        \STATE $\delta_i \gets \text{Proj}_{\epsilon} \left( \delta_i - \eta \cdot \text{sign}(\nabla_{\delta_i} \mathcal{L}_{outer}) \right)$;
        
        \STATE $\theta' \gets \theta'^{(t)}$; \quad \textit{// Reset surrogate model}
    \ENDFOR
\ENDFOR
\STATE $\mathcal{D}_u \gets \{(x_i + \delta_i, y_i) \mid (x_i, y_i) \in \mathcal{D}_c\}$;
\end{algorithmic}
\end{algorithm}
\section{Experiments}
\label{sec:Experiments}
\subsection{Experimental Setups}
\subsubsection{Datasets and Models} 
We evaluate the effectiveness of our method on three benchmark datasets: CIFAR-10~\cite{cifar10}, CIFAR-100~\cite{cifar10}, and Tiny ImageNet~\cite{Tiny-imagenet}. These datasets vary in resolution and complexity, serving as standard benchmarks for image classification.
Regarding the model architecture, following the standard evaluation protocol for unlearnable examples~\cite{EMN, REM}, we primarily use ResNet-18~\cite{resnet18} as the backbone network to simulate the unauthorized model. 
To rigorously assess the resistance of unlearnable examples against transfer learning, we utilize models initialized with weights trained on different datasets.
Additionally, we also conduct experiments using pretrained models ResNet-50~\cite{resnet50}, DenseNet-121~\cite{Densenet121}, and VGG-11~\cite{VGG11} to evaluate the cross-architecture transferability of our method.

\subsubsection{Training Paradigms} 
To rigorously evaluate protection performance, we evaluated our approach across three training paradigms:
training from scratch, standard pretraining–finetuning, and semantic-focused pretraining–finetuning.
1) \textit{Full Training}: As a standard baseline, the unauthorized model is initialized with random weights and trained on the unlearnable dataset. This setting evaluates the basic effectiveness of UEs without prior knowledge interference. 2) \textit{Standard PF}: Based on the vulnerability analysis in Section~\ref{sec:preliminary}, we found that freezing shallow layers maximally reduces unlearnability under the Standard Pretraining–Finetuning paradigm. Specifically, the configuration freezing conv1 and layer1achieves the highest clean test accuracy. To rigorously evaluate the effectiveness of unlearnable examples, we adopt this configuration as the default setting for all subsequent pretraining experiments. 3) \textit{SF-PF}: Building upon the standard PF paradigm, we introduce the semantic-focused Pretraining–Finetuning as a more strict training paradigm. By introducing detection branches and depth supervision into shallow layers during pretraining, SF-Pretrain significantly enhances the pretrained model's ability to filter out semantic mismatch noise.

\setlength{\dbltextfloatsep}{20pt}
\begin{table*}[t]
\centering
\caption{Comparison of clean test accuracy (\%) $\downarrow$ on CIFAR-10, CIFAR-100, and Tiny-ImageNet. Using ResNet-18 as the unauthorized model. Lower accuracy indicates better protection performance. 
}
\label{tab:combined_results_refined}
\setlength{\tabcolsep}{6pt}
\resizebox{\textwidth}{!}{%
\begin{tabular}{c|c|c|cccccccc}
\toprule
\multirow{2}{*}{\bf Dataset} & \multirow{2}{*}{\bf Training Paradigm} & {\bf Pretraining Dataset} & \multicolumn{8}{c}{\bf UEs Generation Method} \\
\cmidrule(l){4-11}
 & & (ResNet-18) & 
 \textbf{EMN} & \textbf{NTGA} & \textbf{REM} & \textbf{AR} & \textbf{SHR} & \textbf{VTG} & \textbf{GUE} & \textbf{Ours} \\
\midrule
\multirow{6}{*}{\bf CIFAR-10} & Full Training & --- & 18.47 & \textbf{11.83} & 17.25 & 13.91 & 12.06 & 16.58 & 13.62 & 12.34 \\
\cmidrule{2-11}
 & \multirow{3}{*}{Standard PF} & ImageNet & 65.19 & 66.77 & 50.42 & 64.05 & 46.88 & 63.51 & 55.29 & \textbf{32.96} \\
 & & Tiny-ImageNet & 53.63 & 52.14 & 32.75 & 62.38 & 33.57 & 54.92 & 54.11 & \textbf{22.68} \\
 & & CIFAR-100 & 57.71 & 61.33 & 52.48 & 53.65 & 49.82 & 55.15 & 59.59 & \textbf{38.27} \\
\cmidrule{2-11}
 & SF-PF & CIFAR-100 & 72.04 & 69.66 & 56.23 & 66.91 & 60.55 & 70.78 & 67.40 & \textbf{47.17} \\
\midrule
\multirow{6}{*}{\bf CIFAR-100} & Full Training & --- & 7.42 & 4.87 & 11.33 & \textbf{4.19} & 8.65 & 8.04 & 9.71 & 5.58 \\
\cmidrule{2-11}
 & \multirow{3}{*}{Standard PF} & ImageNet & 64.27 & 51.95 & 40.66 & 63.48 & 46.13 & 62.81 & 64.55 & \textbf{24.39} \\
 & & Tiny-ImageNet & 56.51 & 42.76 & 32.88 & 56.11 & \textbf{27.35} & 54.62 & 50.97 & \textbf{15.24} \\
 & & CIFAR-10 & 68.09 & 56.44 & 51.70 & 58.53 & 49.88 & 63.26 & 55.61 & \textbf{30.17} \\
\cmidrule{2-11}
 & SF-PF & CIFAR-10 & 72.85 & 62.13 & 62.49 & 63.77 & 60.34 & 68.56 & 66.92 & \textbf{49.05} \\
\midrule
\multirow{6}{*}{\bf Tiny-ImageNet} & Full Training & --- & 8.73 & \textbf{4.26} & 16.51 & 5.89 & 9.14 & 8.67 & 12.05 & 9.92 \\
\cmidrule{2-11}
 & \multirow{3}{*}{Standard PF} & ImageNet & 57.18 & 52.84 & 49.35 & 60.77 & 53.41 & 51.63 & 62.29 & \textbf{31.56} \\
 & & CIFAR-100 & 58.93 & 47.70 & 31.44 & 43.58 & 37.21 & 53.07 & 52.86 & \textbf{26.19} \\
 & & CIFAR-10 & 55.34 & 52.11 & 43.79 & 59.45 & 46.90 & 50.28 & 49.67 & \textbf{27.52} \\
\cmidrule{2-11}
 & SF-PF & CIFAR-10 & 61.88 & 55.46 & 47.03 & 62.12 & 49.57 & 55.71 & 53.24 & \textbf{34.69} \\
\midrule
 \rowcolor{gray!20} 
\multicolumn{3}{c|}{\textbf{Average Accuracy}} & 52.76 & 48.21 & 38.74 & 55.03 & 41.24 & 51.67 & 51.03 & \textbf{28.94} \\
\bottomrule
\end{tabular}%
}
\end{table*}
\vspace{5pt}
\subsubsection{Compared Methods} 
We compare our proposed hierarchical deception strategy with seven state-of-the-art unlearnable examples generation methods, covering distinct generation mechanisms: EMN~\cite{EMN}, the pioneering approach based on the error-minimizing noise paradigm; REM~\cite{REM}, which integrates adversarial training to enhance robustness against data augmentations; NTGA~\cite{NTGA}, an algorithm exploiting Neural Tangent Kernel theory to degrade model generalization; AR~\cite{AR}, which prevents feature extraction by constructing adversarial regimes; SHR~\cite{SHR}, an approach combining synthetic noise with robustness optimization objectives; GUE~\cite{GUE}, which reformulates the attack as a nonzero-sum Stackelberg game solved via an autoencoder-like generator; and VTG~\cite{VTG}, which employs adversarial domain enhancement and perturbation-label coupling to improve transferability across varying architectures and resolutions.
\vspace{5pt}
\subsubsection{Experiment Settings}
We reproduce all the compared methods under the identical experimental settings. For a fair evaluation, all methods, including ours, are restricted to an $\ell_{\infty}$-norm perturbation budget of $\epsilon=8/255$ with respect to the image pixel range $[0, 1]$.
For all experiment settings, we enforce a unified training configuration to ensure fair comparison: models are trained with SGD, momentum 0.9, weight decay $5 \times 10^{-4}$, and a batch size of 128. We train for 100 epochs with a standard learning rate schedule.

\vspace{5pt}
\subsubsection{Evaluation Metrics}
To assess the privacy protection capability of unlearnable examples, we adopt Clean Test Accuracy as the evaluation metric. This metric quantifies classification accuracy of models trained on unlearnable datasets and evaluated on clean test sets. Lower clean test accuracy suggests models fail to learn effective information from training data, indicating stronger privacy protection of UEs.

\subsection{Full Training from Scratch}
We first evaluate the effectiveness of unlearnable examples under the standard full Training from Scratch paradigm, as summarized in Table~\ref{tab:combined_results_refined}. 
Across datasets of increasing complexity, including CIFAR-10, CIFAR-100, and Tiny-ImageNet, our method consistently achieves state-of-the-art protection performance.
On CIFAR-10, our approach restricts clean test accuracy to 12.34\%, performing comparably to strong baselines such as NTGA, which demonstrates competitive performance in low-complexity settings. As dataset complexity increases, the advantage of our method becomes more pronounced. On Tiny-ImageNet, the more complex dataset in our evaluation, which features higher resolution and diverse semantics, our method maintains a low accuracy of 9.92\%. This result highlights the method’s superior ability to inject persistent unlearnable signals, even when the model must learn highly intricate feature representations. These findings demonstrate that our approach scales effectively with the complexity of the dataset and outperforms baseline methods.

\subsection{Robustness under the PF Paradigm}
We next evaluate robustness under the more practical and challenging pretraining–finetuning paradigm, where unauthorized models are initialized from pretrained model weights and shallow layers are frozen. 
In this setting, the performance gap between our method and the compared methods becomes significantly more pronounced.
As shown in Table~\ref{tab:combined_results_refined}, compared methods experience a severe protection collapse under the PF paradigm. 
This degradation can be attributed to frozen, pretrained shallow layers that suppress perturbations lying off the natural image feature manifold, causing unlearnable signals to fail.
In contrast, our method demonstrates strong robustness under the same conditions.
For example, on Tiny-ImageNet using an ImageNet-pretrained ResNet-18, established methods such as AR and GUE fail substantially, with clean test accuracies rising to 60.77\% and 62.29\%, respectively. Meanwhile, our method suppresses the accuracy to 31.56\%, outperforming the strongest compared method REM by 17.79\%. 
This substantial margin highlights the effectiveness of our approach in overcoming semantic filtering induced by frozen pretrained shallow layers.
\begin{table*}[t]
\centering
\caption{Comparison of clean test accuracy(\%) $\downarrow$ under the Standard PF paradigm across different architectures. All network architectures are pretrained on ImageNet. Lower accuracy indicates better protection performance.
}
\label{tab:architecture_transferability}
\setlength{\tabcolsep}{8pt}
\resizebox{0.9\textwidth}{!}{%
\begin{tabular}{c|c|cccccccc}
\toprule
\multirow{2}{*}{\bf Dataset} & {\bf Architectures} & \multicolumn{8}{c}{\bf UEs Generation Method} \\
\cmidrule(l){3-10}
 & (ImageNet) & 
 \textbf{EMN} & \textbf{NTGA} & \textbf{REM} & \textbf{AR} & \textbf{SHR} & \textbf{VTG} & \textbf{GUE} & \textbf{Ours} \\
\midrule
\multirow{4}{*}{\bf CIFAR-10} 
 & ResNet-18   & 65.19 & 66.77 & 50.42 & 64.05 & 46.88 & 63.51 & 55.29 & \textbf{32.96} \\
 & ResNet-50   & 60.45 & 62.89 & 40.12 & 73.56 & 41.03 & 58.27 & 41.90 & \textbf{35.44} \\
 & DenseNet-121 & 52.28 & 37.02 & 54.10 & 53.17 & 46.94 & 50.31 & 43.35 & \textbf{27.95} \\
 & VGG-11      & 68.83 & 64.37 & 66.10 & 80.73 & 50.70 & 65.09 & 59.16 & \textbf{42.17} \\
\midrule
\multirow{4}{*}{\bf CIFAR-100} 
 & ResNet-18   & 64.27 & 51.95 & 40.66 & 63.48 & 46.13 & 62.81 & 64.55 & \textbf{24.39} \\
 & ResNet-50   & 56.78 & 56.02 & 33.45 & 57.67 & 42.29 & 57.90 & 66.14 & \textbf{27.83} \\
 & DenseNet-121 & 41.94 & 51.45 & 41.92 & 45.12 & 41.95 & 42.45 & 67.74 & \textbf{29.37} \\
 & VGG-11      & 67.93 & 49.61 & 64.58 & 61.39 & 35.86 & 64.69 & 67.58 & \textbf{24.41} \\
\midrule
\multirow{4}{*}{\bf Tiny-ImageNet} 
 & ResNet-18   & 57.18 & 52.84 & 49.35 & 60.77 & 53.41 & 51.63 & 62.29 & \textbf{31.56} \\
 & ResNet-50   & 66.47 & 59.02 & 53.88 & 68.15 & 49.74 & 62.39 & 67.61 & \textbf{36.25} \\
 & DenseNet-121 & 52.24 & 58.72 & 49.16 & 48.61 & 43.11 & 54.07 & 64.49 & \textbf{34.72} \\
 & VGG-11      & 58.92 & 56.71 & 48.22 & 58.49 & 50.38 & 57.44 & 67.93 & \textbf{37.65} \\
\midrule
 \rowcolor{gray!20} 
\multicolumn{2}{c|}{\textbf{Average Accuracy}} & 58.54 & 55.61 & 49.33 & 62.19 & 45.95 & 57.55 & 60.67 & \textbf{32.06} \\
\bottomrule
\end{tabular}%
}
\end{table*}
\subsection{Robustness under SF-Pretrain paradigm}
Finally, we subject the proposed approach to a more rigorous training paradigm, SF-Pretrain, which explicitly penalizes semantic mismatch features in shallow layers. 
This setting represents an extreme scenario where the semantic filtering effect is extremely strengthened. 
As summarized in Table~\ref{tab:combined_results_refined}, our approach outperforms the compared methods under the SF-Pretrain paradigm across different datasets.
For better analysis, we conduct experiments using unlearnable examples generated from CIFAR-10 and evaluate them under pretraining paradigms with varying degrees of semantic-focused.
The penalty strength $\lambda \in \{1, 3, 5, 7\}$ determines the degree of semantic regularization imposed on the shallow layers of the pretrained model. Larger values correspond to stronger suppression of semantic mismatch features. This experiment aims to verify that the filtering of perturbations by frozen shallow layers stems from their semantic mismatch with natural images; as semantic-filtering is progressively enhanced, the suppression of these perturbations intensifies.

As shown in the bottom section of Table~\ref{tab:lambda_comparison}, increasing $\lambda$ substantially reduces the effectiveness of most baseline methods. 
Under the strongest semantic constraint ($\lambda = 7$), the protection provided by EMN and NTGA nearly collapses, with clean test accuracies rising to 80.57\% and 77.23\%, respectively. In contrast, our method remains the most resilient, maintaining a suppressed accuracy of 55.29\% even under aggressive semantic enforcement.
These results suggest that, by explicitly aligning perturbations with the natural feature manifold, our method constructs a form of semantic camouflage.
Although increasing the semantic penalty inevitably weakens the unlearnable effect for all methods,  including ours,this degradation is substantially milder for the proposed approach compared to prior methods that rely on semantic mismatch perturbations.
Consequently, even as the semantic filters become more restrictive, our method consistently preserves a stronger protection effect, allowing unlearnability to persist when semantic mismatch perturbations are largely eliminated.
\begin{table}[htbp]
  \centering
  \caption{\textsc{Comparison of clean test accuracy (\%) on CIFAR-10 under different $\lambda$ values under SF pretraining–finetuning paradigm.}}
  \label{tab:lambda_comparison}
  \setlength{\tabcolsep}{0pt} 
  \renewcommand{\arraystretch}{1.3} 
  \small
  
  \begin{tabular*}{0.95\columnwidth}{@{\extracolsep{\fill}}ccccccc}
    \toprule
    \textbf{Metric} & \textbf{EMN} & \textbf{NTGA} & \textbf{REM} & \textbf{AR} & \textbf{SHR} & \textbf{Ours} \\
    \midrule
    $\lambda = 1$ & 66.47 & 68.92 & 54.15 & 63.88 & 53.50 & \textbf{44.95} \\
    $\lambda = 3$ & 72.04 & 69.66 & 56.23 & 66.91 & 60.55 & \textbf{47.17} \\
    $\lambda = 5$ & 74.36 & 75.09 & 63.74 & 72.41 & 67.65 & \textbf{52.40} \\
    $\lambda = 7$ & 80.57 & 77.23 & 66.78 & 73.12 & 69.84 & \textbf{55.29} \\
    \bottomrule
  \end{tabular*}
\end{table}
\subsection{Scalability to Target Dataset Complexity}
To systematically assess the scalability of the proposed method, we evaluate its protection efficacy across datasets characterized by higher resolutions and more intricate semantic spaces.
As detailed in Table~\ref{tab:combined_results_refined}, while baseline methods achieve competitive results on the relatively simple CIFAR-10 dataset, their protection capabilities collapse on the more complex Tiny-ImageNet. For instance, under the standard pretraining paradigm pretrained on Imagenet, methods such as AR and GUE fail substantially on Tiny-ImageNet, yielding clean test accuracies of 60.77\% and 62.29\% respectively. Conversely, our method consistently maintains a robust unlearnable effect, suppressing the accuracy to 31.56\% and outperforming the strongest baseline REM by a margin of 17.79 percentage points. 
This suppression demonstrates that the proposed hierarchical deception strategy SSC scales effectively without being compromised by the rich semantic diversity and high dimensionality of the protected data.
\begin{figure}[t]
    \centering
    \includegraphics[width=\linewidth]
    {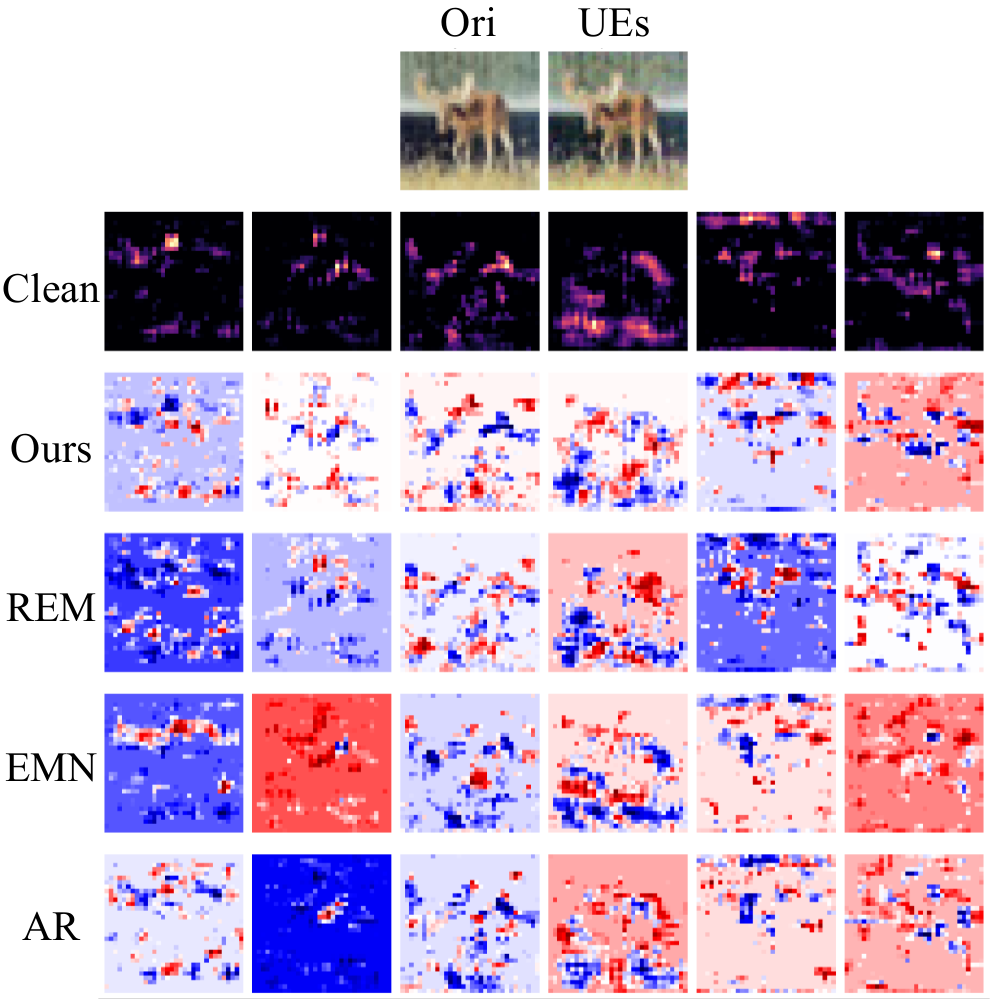}
\caption{Feature perturbation responses in shallow layers of an SF-Pretrain model.
The second row presents the feature activations of the clean example. The following rows visualize feature-level perturbation residuals using the Seismic colormap, where red and blue indicate feature enhancement and suppression, respectively, and white denotes minimal change.}
\label{fig:feature_vis}
\end{figure}

\subsection{Robustness in Black-Box Scenarios}
In practical applications, protectors generally do not possess prior knowledge of the architecture or pretraining data utilized by unauthorized models. Consequently, a practical protection method must exhibit robustness across diverse pretraining datasets and ensure cross-architecture transferability.

\subsubsection{Robustness across Diverse Pretraining Datasets}
Unauthorized models are frequently initialized with weights pretrained on diverse large-scale datasets to accelerate convergence and enhance feature extraction capabilities. To confirm the cross-dataset defense efficacy of our method, we evaluate its resilience against variations in pretraining data distributions. 

As shown in the Pretraining sections of Table~\ref{tab:combined_results_refined}, the proposed approach maintains a formidable unlearnable effect regardless of pretraining data origin. Specifically, when an unauthorized model utilizes weights pretrained on Tiny-ImageNet to learn from unlearnable examples generated on CIFAR-100, our strategy restricts the test accuracy to a mere 15.24\%. Under the identical cross-domain setting, baseline methods including EMN and NTGA fail to preserve their protective capabilities, permitting accuracies to surge to 56.51\% and 42.76\%, respectively. Furthermore, in near-domain transfer scenarios where unlearnable CIFAR-10 examples are evaluated on models initialized with CIFAR-100 weights, our method limits the accuracy to 38.27\%, significantly outperforming the most competitive baseline REM at 52.48\%. These outcomes verify that the proposed hierarchical deception strategy SSC aligns perturbations with natural image feature manifolds, ensuring robustness across diverse pretraining datasets.
\subsubsection{Cross-Architecture Transferability}
Beyond diverse pretraining datasets, unauthorized users may deploy a wide array of neural network architectures varying in depth and capacity. To evaluate cross-architecture transferability, we generate unlearnable examples utilizing a lightweight ResNet-18 model and assess their effectiveness against various unseen network architectures. Crucially, to isolate the impact of architectural variations from pretraining data biases, all evaluated models are pretrained on the same dataset, ImageNet.

The comprehensive results in Table~\ref{tab:architecture_transferability} demonstrate that our method consistently achieves the lowest clean test accuracy across all evaluated architectures and datasets. Notably, the protective efficacy remains highly robust not only on deeper models with expanded capacity like ResNet-50, where it restricts CIFAR-100 accuracy to 27.83\%, but also on architectures employing entirely distinct feature extraction paradigms. When confronting the dense connectivity pattern of DenseNet-121, our method successfully suppresses the accuracy to 27.95\% on CIFAR-10 and 34.72\% on Tiny-ImageNet. Similarly, against the traditional sequential convolutions of VGG-11, the accuracy is firmly held at 24.41\% on CIFAR-100. These findings provide compelling evidence that the increased parameter capacity and varied structural designs of different models are insufficient to bypass the deeply integrated unlearnable perturbations introduced by our approach.

\subsection{Semantic Perception Analysis in Shallow Layers}
We qualitatively analyze the perceptual representations of different UEs generation methods by extracting features from the first layer of the SF-Pretrain model. 
By extracting feature perturbations from the initial layers, we analyze whether the model's perceptual focus on these noises aligns with the intrinsic image semantics. To isolate the effect of perturbations from the underlying image structure, we compute feature-level residuals induced solely by the injected noise. These residuals are visualized using the Seismic colormap, where red denotes feature enhancement, blue indicates suppression, and white corresponds to negligible changes.

As shown in Figure~\ref{fig:feature_vis}, the second row presents the standard activations of clean samples, serving as a reference for evaluating perturbation patterns. 
Compared with conventional baseline methods, our channel-wise semantic perturbations produce responses with lower intensity in the residual maps in non-semantic regions, as reflected by reduced color saturation. Such moderate feature deviations avoid inducing extreme activations, which may otherwise be attenuated by frozen shallow layers in pretrained models.
Moreover, while baseline methods tend to activate spatially irrelevant regions in certain channels, our perturbations exhibit strong spatial alignment with the corresponding clean feature maps. The activation regions closely follow the semantic boundaries of target objects, indicating a higher degree of semantic consistency.

These observations demonstrate that our method generates perturbations whose feature responses align well with the original image semantics across channels. As a result, the perturbations resemble natural feature variations rather than anomalous signals, allowing them to propagate more effectively through frozen shallow layers and influence deeper representations during subsequent training processing.
\section{CONCLUSION}
In this study, we conducted the first systematic analysis of UEs under the pretraining–finetuning paradigm. We identified a critical vulnerability in which standard UEs, which depend on shallow shortcuts, are suppressed by the frozen shallow layers of pretrained models as a result of a semantic filtering effect.
To address this vulnerability, we introduce SSC, a method that aligns perturbations with shallow semantic representations to circumvent semantic filtering and concentrate the poisoning effect in deep, trainable layers. Comprehensive experiments on CIFAR-10, CIFAR-100, and Tiny-ImageNet show that SSC consistently surpasses existing approaches, preserving the efficacy of unlearnable examples across diverse training paradigms.
In summary, this work reveals a previously unrecognized vulnerability of UEs in contemporary transfer learning pipelines and offers a practical approach for safeguarding data privacy across diverse training paradigms.

\bibliographystyle{IEEEtran}
\bibliography{ref}

\end{document}